\documentclass[sigconf,nonacm]{acmart}
\setcopyright{none}

\AtBeginDocument{%
  }

\usepackage[table]{xcolor}
\usepackage{algorithm}
\usepackage{multicol}
\usepackage{array} 
\usepackage{multirow}
\usepackage{makecell}
\usepackage{graphicx}
\usepackage{subcaption}
\usepackage{enumitem}

\begin{document}

\title{Dynamic Thermal Gaussians: Multimodal 4D Gaussian Splatting}

\author{Rongfeng Lu}
\authornote{Both authors contributed equally to this research.}
\affiliation{%
  \institution{Lishui University}
  \city{Lishui}
  \country{China}
}
\affiliation{%
  \institution{Lishui Key Laboratory of Low-Altitude Multimodal Sensing and Intelligent Computing}
  \city{Lishui}
  \country{China}
}
\email{rongfeng-lu@hdu.edu.cn}
\orcid{0009-0002-8002-4688}

\author{Lifeng Lin}
\authornotemark[1]
\affiliation{%
  \institution{Hangzhou Dianzi University}
  \city{Hangzhou}
  \country{China}
}
\email{lifeng\_lin@hdu.edu.cn}
\orcid{0009-0005-3150-1857}

\author{Xiaobao Wei}
\authornotemark[1]
\affiliation{%
  \institution{University of Chinese Academy of Sciences}
  \city{Beijing}
  \country{China}
}
\email{weixiaobao0210@gmail.com}
\orcid{0000-0003-4230-1162}

\author{Quan Chen}
\affiliation{%
  \institution{Jiaxing University}
  \city{Jiaxing}
  \country{China}
}
\email{chenquan@zjxu.edu.cn}
\orcid{0000-0003-2858-6771}

\author{Ming Lu}
\affiliation{%
  \institution{Hangzhou Dianzi University}
  \city{Hangzhou}
  \country{China}
}
\email{lu199192@gmail.com}
\orcid{0000-0001-6819-6490}

\author{Yitian Xue}
\affiliation{%
  \institution{Zhejiang University}
  \city{Hangzhou}
  \country{China}
}
\email{yitianxue@zju.edu.cn}
\orcid{0009-0003-3623-0422}

\author{Yaoqi Sun}
\correspondingauthor
\authornote{Corresponding author.}
\affiliation{%
  \institution{Lishui University}
  \city{Lishui}
  \country{China}
}
\email{sunyq2233@163.com}
\orcid{0000-0001-8874-241X}

\author{Yuhan Gao}
\affiliation{%
  \institution{Hangzhou Dianzi University}
  \city{Hangzhou}
  \country{China}
}
\email{yuhangao@hdu.edu.cn}
\orcid{0000-0002-7777-3613}

\author{Anke Xue}
\affiliation{%
  \institution{Hangzhou Dianzi University}
  \city{Hangzhou}
  \country{China}
}
\email{akxue@hdu.edu.cn}
\orcid{0000-0001-8313-8520}

\author{Chenggang Yan}
\affiliation{%
  \institution{Hangzhou Dianzi University}
  \city{Hangzhou}
  \country{China}
}
\email{cgyan@hdu.edu.cn}
\orcid{0000-0003-1204-0512}

\renewcommand{\shortauthors}{Rongfeng Lu et al.}

\begin{abstract}
Thermography plays a vital role in military and broader thermal analysis applications. Recent progress in 3D thermal reconstruction has extended temperature analysis from 2D to 3D space, yet most existing works assume static temperature distributions, neglecting the temporal dynamics of heat transfer in real-world environments. To address this limitation, we propose the first dynamic RGB-Thermal reconstruction framework for complex scenes. Our method jointly models RGB appearance, thermal observations, and scene geometry as they change over time. Specifically, we introduce a multimodal dynamic scene representation that anchors both the color and thermal modalities to a shared geometric substrate, ensuring their consistency under spatiotemporal deformations. We further design multimodal embeddings to enhance the motion expressiveness for each modality, and propose a multimodal routing mechanism that retains a unified set of shared multimodal Gaussians as the geometric backbone while adaptively spawning modality-specific Gaussians to strengthen the representational capacity in detail-rich regions of each individual modality. In addition, we contribute a novel benchmark dataset featuring high-frequency temperature variations to facilitate the evaluation of 4D reconstruction. Extensive experiments demonstrate that our method achieves high-fidelity spatiotemporal reconstruction of both appearance and temperature. Our code and dataset are available at: https://github.com/LinLif1869/DTG.
\end{abstract}

\keywords{Multimodal fusion, 4D scene reconstruction, Novel view synthesis, 3D Gaussian splatting}

\maketitle
\section{Introduction}
\label{sec: intro}

Temperature governs diverse natural and engineered processes. Thermal imaging converts surface temperature distributions into interpretable 2D observations and supports applications ranging from military surveillance~\cite{he2021infrared} and industrial analysis~\cite{glowacz2021fault} to architecture~\cite{el2020scoping}, search and rescue~\cite{yeom2024thermal}, and medical diagnostics~\cite{lahiri2012medical}. In real-world environments, thermal fields evolve continuously and can change rapidly under convection, conduction, and radiation.
Existing 3D thermal field reconstruction methods characterize spatial geometry but do not effectively capture complex dynamic heat transfer. Therefore, extending thermal field reconstruction into the 4D spatiotemporal domain has become key to advancing this field.

Recent advances in Neural Radiance Fields~(NeRF) \cite{mildenhall2021nerf} and 3D Gaussian Splatting~(3DGS)~\cite{kerbl20233d} have significantly improved high-fidelity novel view synthesis and 3D reconstruction. Building on them, numerous studies~\cite{hassan2024thermonerf, lin2024thermalnerf, luthermalgaussian, chen2024thermal3d} target high-quality static 3D thermal field reconstruction. Recent methods further model temporal thermal evolution over RGB-derived or fixed scene geometry~\cite{thermalgs2025,yang2025ntr,wang2026etgs}. They do not jointly reconstruct dynamic RGB appearance, thermal observations, and scene geometry in complex scenes with object motion and rapid temperature changes.
We address this setting with the first dynamic RGB-Thermal reconstruction framework for complex scenes.

The lack of open-source thermal field reconstruction datasets has constrained progress in 4D thermal reconstruction. Existing benchmarks do not cover joint RGB-Thermal reconstruction in complex scenes where geometry and temperature change rapidly. To support research in this setting, we introduce \textbf{DynamicRGBT-Scenes}, a benchmark dataset for dynamic RGB-Thermal scene reconstruction with fast temperature changes.
DynamicRGBT-Scenes contains 12 synchronized dual-view RGB-Thermal sequences captured using two handheld device pairs, each consisting of a FLIR A700 thermal camera and an iPhone 15 Pro. The scenes include open-flame grilling, electric iron soldering, hair drying, and clothes ironing, among others. They cover rapid temperature changes, object motion, low light, and occlusion.

Representative 4D RGB methods include 4DGaussians~\cite{wu20244d}, DN-4DGS~\cite{lu2024dn}, and E-D3DGS~\cite{bae2024per}. Recent work further extends dynamic rendering to manipulation~\cite{li2025manipdreamer3d}, robotic~\cite{wang2025roboarmgs}, and street scenes~\cite{wei2025emd}. The discrete 3DGS representation provides flexibility for complex topological changes. Extending these methods to thermal fields remains difficult because thermal images often lack texture and distinctive features, leading to unstable optimization.
To address this challenge, we propose a multimodal dynamic reconstruction framework that integrates thermal and RGB modalities. We build a unified Gaussian representation that preserves geometric consistency and temporal continuity, where each Gaussian encodes both color and thermal attributes.
We further introduce multimodal embeddings and dynamic deformation to model temporal evolution, along with a multimodal routing mechanism that separates shared and modality-specific Gaussians to enhance detail representation.
Comprehensive experiments show that our method achieves state-of-the-art dynamic thermal field reconstruction.

In summary, the main contributions are as follows:

(1) We present \textbf{DynamicRGBT-Scenes}, the first benchmark to capture rapid thermal variations in dynamic environments with synchronized RGB–thermal image pairs from dual viewpoints.

(2) We propose \textbf{Dynamic Thermal Gaussians}, the first dynamic RGB-Thermal reconstruction framework for complex scenes, rendering high-fidelity RGB and thermal views at any timestep.

(3) We design multimodal embeddings and a progressive deformation architecture to model the distinct temporal evolution of each modality. We further integrate a learnable multimodal routing mechanism into the unified dynamic Gaussian representation to automatically distinguish shared and modality-specific Gaussians. 

(4) Extensive experiments demonstrate that our multimodal method achieves state-of-the-art performance across both thermal and RGB modalities, establishing a solid foundation for spatiotemporal thermography.

\section{Related Work}
\label{sec: relwork}

\subsection{Thermal Scene Reconstruction}

Thermal imaging estimates surface temperature from infrared radiation emitted by objects, typically within the mid-wave or long-wave infrared spectrum. The resulting temperature distributions are commonly visualized using pseudo-color mapping. Since thermal measurements do not depend on visible illumination, thermal imaging can provide informative observations in low-light environments and under partial occlusion.

Multi-view thermal images enable 3D temperature analysis beyond fixed viewpoints. KinectFusion~\cite{newcombe2011kinectfusion} established dense surface mapping, and later systems added infrared observations~\cite{rangel20143d,muller2019close,li2023research}. These static methods lack photorealistic novel-view synthesis and adapt poorly across thermal conditions. ThermoNeRF \cite{hassan2024thermonerf}, Thermal-NeRF \cite{ye2024thermal}, and ThermalNeRF \cite{lin2024thermalnerf} extend NeRF to static RGB-Thermal or thermal-only fields, but costly implicit representations limit real-time rendering and structural control.

To address these limitations, recent work adopts 3DGS as a more efficient and explicit alternative. ThermalGaussian and ThermalGaussian++~\cite{luthermalgaussian,lu2026tg} align and render RGB and thermal views; Thermal3D-GS \cite{chen2024thermal3d} adds atmospheric and conductive priors for physical consistency; Veta-GS \cite{nam2025veta} uses view-dependent deformation to improve thermal edges; and MMOne \cite{gu2025mmone} unifies multimodal inputs in 3DGS.

Beyond static reconstruction, dynamic thermal reconstruction has also received increasing attention. ThermalGS~\cite{thermalgs2025} models temporal thermal radiation with Gaussian geometry guided by an RGB mesh and semantic features. NTR-Gaussian~\cite{yang2025ntr} predicts nighttime thermal evolution from learned thermodynamic parameters and numerical integration. ETGS~\cite{wang2026etgs} assigns explicit thermal parameters to Gaussians and uses closed-form heat transfer to predict arbitrary timestamps. These methods model thermal evolution over RGB-derived or fixed geometry, whereas we jointly reconstruct time-varying RGB appearance, thermal observations, and scene geometry.

\subsection{Dynamic Scene Rendering}

Early dynamic NeRF methods extend spatial representations over time~\cite{wei2024nto3d}. D-NeRF~\cite{pumarola2021d} conditions a deformation field on time to model non-rigid motion. Nerfies~\cite{park2021nerfies} and HyperNeRF~\cite{park2021hypernerf} use per-frame latent codes for canonical-space deformation, with HyperNeRF lifting the representation to higher dimensions for topology changes. HyperReel~\cite{attal2023hyperreel} combines ray-conditioned sampling with a compact dynamic volume for efficient, high-quality 6-DoF video synthesis.

Beyond NeRF-based methods, explicit 3DGS representations support generalizable~\cite{wu2025textsplat} and sparse-view~\cite{lin2025vgnc} reconstruction, while their 4D extensions provide an efficient alternative to NeRF volume rendering~\cite{huang2024s3g, wei2025gazegaussian, wei2026parkgaussian}. D-3DGS~\cite{yang2024deformable} learns canonical time-dependent deformation for high-fidelity dynamic reconstruction. 4DGaussians~\cite{wu20244d} combine spatiotemporal neural voxels with Gaussians for faster rendering with explicit structure. E-D3DGS~\cite{bae2024per} uses per-Gaussian embeddings to drive dual-level deformation for complex motion and detail. Other methods use structured 4D Gaussians~\cite{yang2023real}, denoised spatiotemporal aggregation~\cite{lu2024dn}, and lightweight predictors with entropy constraints~\cite{zhang2025mega} to improve appearance, stability, and storage, respectively.

These methods reconstruct dynamic geometry and appearance from RGB observations but omit thermal measurements. Dynamic Thermal Gaussians instead integrates RGB and thermal supervision in one dynamic Gaussian representation for complex scenes.

\begin{figure*}[htbp!]
	\centering
	\includegraphics[scale=0.38]{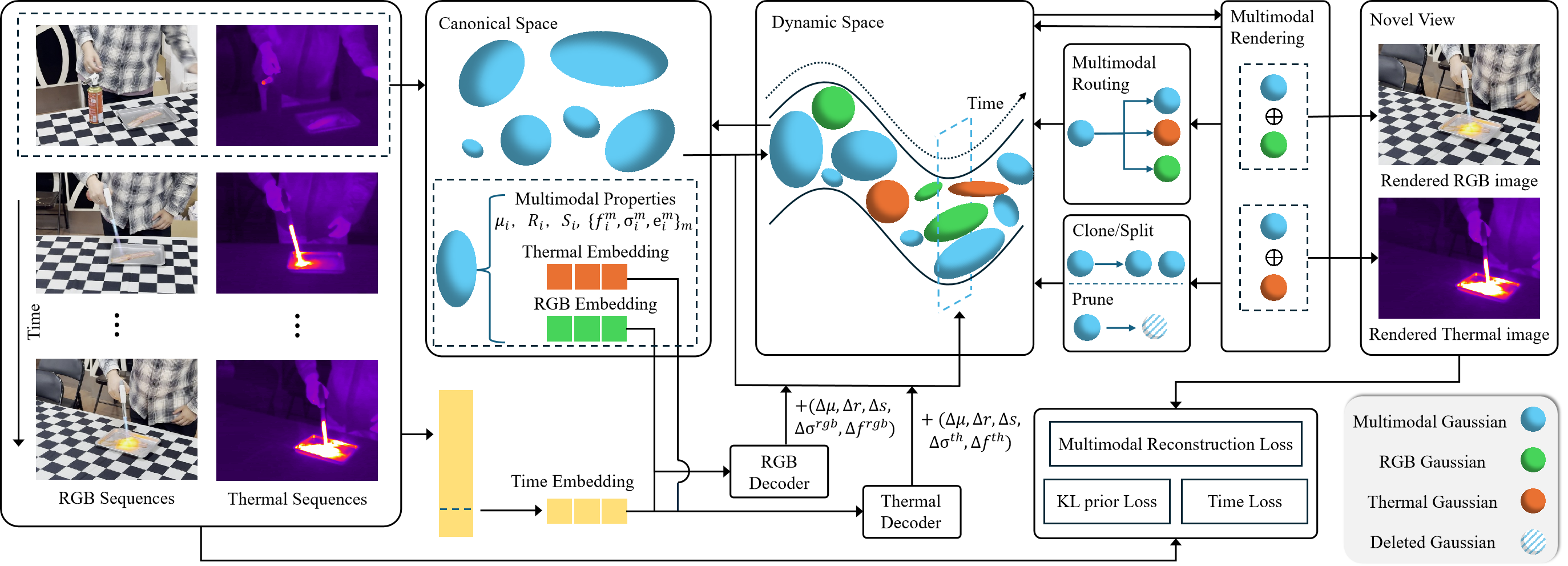}
	\caption{\textbf{Overview.} We first construct a geometrically consistent multimodal dynamic representation in a shared canonical space, then model temporal evolution through embedding-based multimodal deformation, and finally employ a multimodal routing mechanism to adaptively spawn modality-specific Gaussians that enhance detail representation for each modality.
    }
  \Description{pipeline}
	\label{fig:pipeline}
\end{figure*}

\section{Method}
\label{sec: label}
As illustrated in Figure~\ref{fig:pipeline}, we construct Dynamic Thermal Gaussians, a unified framework for multimodal dynamic reconstruction. \textbf{Multimodal Dynamic Modeling} (Section~\ref{subsec:MMM}) combines shared canonical Gaussian geometry with modality-specific attributes; \textbf{Multimodal Embedding-Based Deformation} (Section~\ref{subsec:MED}) models dynamic variations using temporal and modality-specific Gaussian embeddings; and \textbf{Multimodal Routing Mechanism} (Section~\ref{subsec:MRM}) adaptively assigns Gaussians to different modality roles.

\subsection{Preliminary}
\label{subsec: 3DGS}
3D Gaussian Splatting (3DGS)~\cite{kerbl20233d} is an explicit representation for 3D reconstruction and rendering that models a scene with anisotropic 3D Gaussians. Each Gaussian jointly represents geometry, opacity, and view-dependent appearance. The fully differentiable formulation supports end-to-end optimization, while splatting provides efficient rasterization-style rendering.

The 3DGS pipeline starts from a collection of unordered input images, which are processed through Structure-from-Motion (SfM)~\cite{schoenberger2016sfm} to estimate camera poses and generate a sparse point cloud. These sparse points are then used to initialize the Gaussian centers $\mu$. Each Gaussian is represented as:
\begin{equation}
    G(x)=e^{-\frac{1}{2}(x-\mu)^{T} \Sigma^{-1}(x-\mu)},
\end{equation}
where $\Sigma$ represents the covariance matrix, and $x$ denotes any point in the 3D scene.
Each 3D Gaussian is further projected onto the image plane through the camera's intrinsic matrix, yielding a corresponding 2D Gaussian distribution. The rendered image is obtained by alpha compositing:
\begin{equation}
    C\left(x^{\prime}\right)=\sum_{k \in Q} c_{k} \alpha_{k} \prod_{j=1}^{k-1}\left(1-\alpha_{j}\right),
\end{equation}
where $x^{\prime}$ denotes the target pixel, $Q$ is the set of Gaussians contributing to that pixel. $\alpha_k$ represents the opacity of the $k$-th Gaussian, and $c_k$ is its color represented by spherical harmonics. All Gaussian attributes are learnable and optimized directly in an end-to-end manner during training.

\subsection{Multimodal Dynamic Modeling}  
\label{subsec:MMM}

To jointly model multimodal dynamic observations (e.g., RGB and thermal) within a unified 3D Gaussian representation while capturing modality-specific temporal imaging variations, we introduce a multimodal dynamic Gaussian representation. This formulation explicitly encodes modality-specific attributes into the Gaussian representation under the premise of maintaining multimodal geometric consistency and temporal continuity. It serves as a unified representational foundation for subsequent multimodal deformation modeling and training.

In dynamic multimodal scenes, different sensing modalities may exhibit substantially different imaging mechanisms. However, at any given timestamp, they observe the same underlying physical scene. 
Therefore, we design a shared geometric representation for all modalities. Each 3D Gaussian $G_{i}$ is parameterized by its mean position $\mu_{i}$ and covariance $\Sigma_i$, represented via rotation $R_i$ and scale $S_i$. These geometric parameters are shared across all modalities and serve as a common reference coordinate system for time-dependent deformation. This shared canonical space enforces spatial alignment between modalities and ensures that temporal updates remain geometrically consistent.

Based on shared geometry, we introduce modality-specific attributes to capture cross-modal appearance differences. For each modality $m\in \left \{rgb, thermal \right \}$, each Gaussian $G_{i}$ is associated with: appearance features $f_i^m$, encoding modality-specific radiance; opacity $\sigma_i^m$, modeling modality-dependent visibility; modality embedding vectors $e_i^m$, providing modality-aware conditioning signals for the deformation network.
Unlike ThermalGaussian~\cite{luthermalgaussian}, this representation does not assume equal relevance of each Gaussian across modalities. Instead, modality-dependent participation masks are learned to selectively gate Gaussians while keeping the underlying canonical geometry shared.
Combining shared geometry and modality-specific attributes, each Gaussian can be expressed as
\begin{equation}
    G_{i} = \left(\mu_{i}, R_{i}, S_{i}, \left \{ f_{i}^{m}, \sigma_{i}^{m}, e_{i}^{m} \right \}_{m}\right),
\end{equation}
where geometric parameters $(\mu_{i}, R_{i}, S_{i})$ are dynamically updated by a modality-aware deformation network, and modality-specific attributes capture distinct temporal patterns reflecting the imaging characteristics of each modality.

Overall, this representation models multimodal observations on shared geometry while separating modality-dependent variations into dedicated attribute channels. Decoupling shared geometry from modality-specific attributes provides a stable foundation for dynamic deformation and joint optimization without prematurely assigning each Gaussian to a modality, preserving the flexibility required for adaptive routing.

\subsection{Multimodal Embedding-Based Deformation}
\label{subsec:MED}

Gaussian deformation fields have been shown to be effective for modeling dynamic scenes~\cite{yang2024deformable,wu20244d,yang2023real,bae2024per}. In this formulation, temporal variations are modeled as continuous transformations in the canonical Gaussian space, rather than constructing independent representations for each timestamp, which enforces temporal coherence and improves modeling stability.
However, existing approaches predominantly focus on single-modality observations and do not explicitly account for the heterogeneous deformation behaviors induced by different sensing modalities.

 To address this, we propose a multimodal embedding-based deformation framework, where deformations are jointly conditioned on time embeddings and modality-specific Gaussian embeddings. We further introduce a progressive, modality-separated deformation architecture to capture modality-dependent dynamics under shared geometric constraints.
We condition the deformation field on two complementary types of embeddings: time embeddings and modality-specific Gaussian embeddings. The time embeddings encode the temporal evolution in dynamic scenes and serve as a shared global condition across modalities. For each timestamp $t$, every Gaussian is associated with a time embedding $e_{t} \in \mathbb{R}^{128}$ to guide its continuous temporal evolution.

Building upon E-D3DGS~\cite{bae2024per}, we further construct modality-specific Gaussian embeddings for each Gaussian. As in Section~\ref{subsec:MMM}, every Gaussian $G_{i}$ is associated with a modality-specific embedding $e_{i}^{m} \in \mathbb{R}^{32}$, which encodes its modality-dependent state and provides conditioning signals for deformation.
We model dynamic variations as residual updates over a canonical Gaussian representation, instead of predicting time-dependent absolute parameters.
Given a time embedding $e_{t}$ and a modality-specific Gaussian embedding $e_{i}^{m}$, the deformation network predicts parameter offsets that are additively applied to the static Gaussian parameters.

To explicitly account for the modality-specific roles in dynamic modeling, we adopt a progressive deformation strategy decomposed into two stages. In the first stage, the deformation function $\mathcal{F}^{\text{rgb}} $ takes the time embedding $e_{t}$ and the RGB Gaussian embeddings $e_{i}^{\mathrm{rgb}}$ as input, and predicts residual updates for geometric attributes, RGB appearance, and RGB-specific opacity:
\begin{equation}
\mathcal{F}^{\mathrm{rgb}}:\left(\mathbf{e}_{t}, \mathbf{e}_{i}^{\mathrm{rgb}}\right) \rightarrow\left(\Delta \mathbf{\mu}, \Delta \mathbf{r}, \Delta \mathbf{s}, \Delta \sigma^{\mathrm{rgb}}, \Delta \mathbf{f}^{\mathrm{rgb}}\right),
\end{equation}
where $\Delta(\cdot)$ denotes residuals w.r.t. the canonical parameters. This stage captures the dominant structural motion and the dynamic appearance variations in the visible modality. Conditioned on the geometry updated in the first stage, the second deformation field is dominated by the thermal modality, further updating geometric attributes, thermal appearance and thermal-specific opacity, while reducing interference with RGB representations during thermal dynamic modeling: 
\begin{equation}
    \mathcal{F}^{\mathrm{th}}:\left(\mathbf{e}_{t}, \mathbf{e}_{i}^{\mathrm{th}}\right) \rightarrow\left(\Delta \mathbf{\mu}, \Delta \mathbf{r}, \Delta \mathbf{s},   \Delta \sigma^{\mathrm{th}}, \Delta \mathbf{f}^{\mathrm{th}}\right).
\end{equation}
The thermal reconstruction loss is backpropagated through $\mathcal{F}^{\mathrm{th}}$ to the shared geometric parameters as well as the thermal appearance and opacity. Thermal observations therefore directly supervise dynamic geometry and cross-modal alignment.

In summary, temporal and modality-specific Gaussian embeddings jointly condition deformation in the canonical Gaussian space. Combined with residual parameterization and a progressive modality-aware deformation strategy, this design models multimodal dynamics while preserving modality-specific deformation paths, reducing cross-modal interference, and improving modeling stability. It thus provides a unified dynamic representation for subsequent modality-aware training and optimization.

\subsection{Multimodal Routing Mechanism}
\label{subsec:MRM}
Building upon the unified geometry and modality-specific attributes, a remaining key question is determining which regions should be shared across modalities and which should be dominated by a single modality within a unified dynamic Gaussian representation. Forcing every Gaussian to participate in all modalities may introduce unnecessary coupling, while fully separating geometry would break spatial consistency and temporal continuity. To address this, we introduce a multimodal routing mechanism in the shared canonical space, enabling Gaussians to form shared and modality-specific structures adaptively under unified geometric constraints.

Specifically, each Gaussian $G_i$ is associated with a routing logit variable: $\boldsymbol{\ell}_i \in \mathbb{R}^3$, corresponding to three structural roles: \{\text{shared}, \text{rgb-only}, \text{thermal-only}\}.
This variable captures the structural preference of each Gaussian in the multimodal representation.
Since structural roles are discrete, directly applying hard assignment would make the optimization non-differentiable. We therefore adopt the Gumbel-Softmax relaxation~\cite{jang2017categorical}. For each structural role $k$: 
\begin{equation}
g_{i,k} = -\log(-\log(u_{i,k})), \quad u_{i,k} \sim \mathcal{U}(0,1),
\end{equation}
\begin{equation}
s_{i,k}^{\mathrm{soft}} =
\frac{
\exp\left((\ell_{i,k} + g_{i,k}) / \tau\right)
}{
\sum_{j} \exp\left((\ell_{i,j} + g_{i,j}) / \tau\right)
}.
\end{equation}

$\tau$ denotes the temperature parameter. The soft assignment $s_{i,k}^{\mathrm{soft}}$ provides a differentiable approximation for gradient propagation. During forward rendering, we use the corresponding hard assignment $s_{i,k}^{\mathrm{hard}}$ and backpropagate gradients with the Straight-Through estimator~\cite{liu2022nonuniform}, enabling stable optimization of the discrete routing variables:
\begin{equation}
s_{i,k}^{\mathrm{hard}} = \mathrm{one\_hot}\left(\arg\max_k s_{i,k}^{\mathrm{soft}}\right).
\end{equation}

The three structural states are further mapped to modality participation masks:
\begin{equation}
\mathrm{Mask}_i^{\mathrm{rgb}} = s_{i,\mathrm{shared}}^{\mathrm{hard}} + s_{i,\mathrm{rgb}}^{\mathrm{hard}},
\quad
\mathrm{Mask}_i^{\mathrm{th}} = s_{i,\mathrm{shared}}^{\mathrm{hard}} + s_{i,\mathrm{thermal}}^{\mathrm{hard}},
\end{equation}
where $\mathrm{Mask}_i^{m} \in \{0,1\}$ indicates whether Gaussian $G_i$ participates in modality $m$. Combined with the modality-specific opacity $\sigma_i^{m}$, the effective opacity becomes:
\begin{equation}
\tilde{\sigma}_i^{m}(t) = \sigma_i^{m}(t) \cdot \mathrm{Mask}_i^{m}.
\end{equation}

Therefore, the modality routing mechanism only regulates the participation contributions of Gaussians across modalities, without altering geometric evolution or multimodal deformation.

Because modality routing introduces discrete structural variables, allowing unrestricted specialization at early training stages may destabilize geometric convergence. We therefore adopt a progressive training strategy. In the initial stage, all Gaussians are treated as shared to ensure stable geometry learning and dynamic deformation modeling. As training proceeds, learnable routing variables are gradually activated, allowing modality preferences to emerge under reconstruction signals. In the specialization stage, routing assignments converge to stable shared and modality-specific regions under the joint influence of reconstruction objectives and global distribution constraints. In the final stage, routing assignments are fixed to ensure stable optimization and evaluation.

Based on this design, we further extend the densification and pruning mechanism of 3DGS~\cite{kerbl20233d} in a structure-consistent manner.
While the original densification and pruning operations rely solely on geometric gradients and opacity, we apply them separately for different structural roles during the specialization stage. Shared Gaussians follow the shared criteria, while rgb-only and thermal-only Gaussians are densified or pruned according to gradients computed from their respective modalities. Newly generated Gaussians inherit the structural role of their parent, preserving structural consistency.

This structure-consistent densification mechanism ensures that modality specialization is reflected not only in participation masks but also in spatial resolution allocation. RGB-dominant regions naturally acquire more rgb-only Gaussians, thermal-dominant regions accumulate more thermal-only Gaussians, and modality-consistent regions remain shared. As a result, routing variables correspond to explicit spatial structures rather than abstract probability weights, enhancing the interpretability of multimodal specialization.

In summary, by introducing structure-aware modality routing into a unified dynamic Gaussian representation and combining it with progressive learning and a structure-consistent densification mechanism, the model adaptively forms shared and modality-specific regions while preserving geometric consistency and temporal continuity, thereby improving the expressiveness and interpretability of dynamic thermal field reconstruction.

We optimize the RGB and thermal renderings with a multimodal reconstruction loss $\mathcal{L}_{\text{rec}}$ that compares each rendered modality with its observation. A routing prior prevents premature structural specialization during early training. Together with the embedding and temporal smoothness terms inherited from E-D3DGS~\cite{bae2024per}, the overall objective is defined as:
\begin{equation}
\mathcal{L}_{\text{total}}=\mathcal{L_{\text{rec}}}+\lambda_{\text{KL}}\mathcal{L}_{\text {KL}}+\lambda_{\text{embed}}\mathcal{L}_{\text {embed}}+\lambda_{\text{time}}\mathcal{L}_{\text {time}}.
\end{equation}

We use RGB $\ell_1$/SSIM and thermal $\ell_1$ reconstruction terms. For routing, $\mathcal{L}_{\text{KL}}=KL(\bar{\mathbf{s}}\|p_0)$ regularizes the mean soft assignment $\bar{s}_k=\frac{1}{N}\sum_i s_{i,k}^{\text{soft}}$ with $\lambda_{\text{KL}}=0.01$. The prior $p_0$ starts from $[0.8,0.1,0.1]$, is set to $[0.5,0.3,0.2]$ when specialization begins, and is linearly annealed to $[0.15,0.7,0.15]$ before the final routing stage.

\section{Self-collected Multimodal Dataset}
\label{sec:dataset}
We collect \textbf{DynamicRGBT-Scenes} with two custom handheld rigs, each equipped with an iPhone 15 Pro and a FLIR A700 thermal camera. All sequences are recorded under smooth camera motion. The commercial-grade FLIR A700 records thermal images at up to \(640 \times 480\) and \(30\,\mathrm{Hz}\), with a measurable range of \(-20\,^{\circ}\mathrm{C}\) to \(120\,^{\circ}\mathrm{C}\), a \(24^{\circ} \times 18^{\circ}\) FOV, and an accuracy of \(\pm 2\%\). The iPhone 15 Pro records \(2\mathrm{K}\) RGB videos at \(30\,\mathrm{Hz}\), which are cropped and downsampled to \(640 \times 480\) for resolution-consistent paired processing.

The dataset contains 12 long-duration sequences spanning temperature variation, object motion, occlusion, and low-light conditions. We manually synchronize device frames and subsample them into long paired RGB-Thermal sequences, aligning both modalities in viewpoint and temporal sampling within hardware and sensor constraints. Following Nerfies~\cite{park2021nerfies}, we reconstruct sparse point clouds and camera poses from RGB frames using COLMAP~\cite{schoenberger2016sfm} and assign them to synchronized thermal frames, providing a stable shared geometric reference for both modalities and for consistent quantitative and qualitative comparisons across methods.

\begin{table*}[!t]
    \centering
    \small
    \caption{Per-scene PSNR comparison. ThermalGaussian~\cite{luthermalgaussian} is shortened as “T-GS”; average SSIM/LPIPS are in Table~\ref{tab2:average_quant}.}
    \resizebox{\textwidth}{!}{
    \begin{tabular}{ccc|cccccccccccc|c}
    \toprule[2pt]
    Modality & Metric & Method
    & \multicolumn{1}{>{\centering\arraybackslash}m{1.0cm}}{\makecell{Heating\\Table}}
    & \multicolumn{1}{>{\centering\arraybackslash}m{1.0cm}}{HotBar}
    & \multicolumn{1}{>{\centering\arraybackslash}m{1.0cm}}{HeatGun}
    & \multicolumn{1}{>{\centering\arraybackslash}m{1.0cm}}{Bacon}
    & \multicolumn{1}{>{\centering\arraybackslash}m{1.0cm}}{IcePacks}
    & \multicolumn{1}{>{\centering\arraybackslash}m{1.0cm}}{HairDryer}
    & \multicolumn{1}{>{\centering\arraybackslash}m{1.0cm}}{\makecell{HairDryer\\Dark}}
    & \multicolumn{1}{>{\centering\arraybackslash}m{1.0cm}}{\makecell{Ironing\\Cloth}}
    & \multicolumn{1}{>{\centering\arraybackslash}m{1.0cm}}{Candles}
    & \multicolumn{1}{>{\centering\arraybackslash}m{1.0cm}}{HotWater}
    & \multicolumn{1}{>{\centering\arraybackslash}m{1.0cm}}{Foam}
    & \multicolumn{1}{>{\centering\arraybackslash}m{1.0cm}}{Covers}
    & \multicolumn{1}{|c}{Avg.} \\
    \midrule[1pt]
        \multirow{6}{*}{Thermal} & \multirow{6}{*}{PSNR$\uparrow$} & D-3DGS \cite{yang2024deformable} & 18.13 & 28.31 & 23.80 & 22.57 & 29.09 & 24.34 & 23.17 & 28.33 & 34.02 & 29.54 & 24.90 & 21.57 & 25.65 \\
        ~ & ~ & 4DGaussians \cite{wu20244d} & 24.83 & 31.28 & 30.00 & 26.60 & 32.33& 27.39 & 29.49 & 30.82 & 33.17 & 30.98 & 36.64 & 29.33 & 30.24 \\
        ~ & ~ & E-D3DGS \cite{bae2024per} & 25.81 & 30.93 & 30.56 & 29.46 & 31.01 & 24.50 & 30.81 & 32.27 & 32.20 & 31.56 & 38.11 & 31.43 & 30.72 \\
        ~ & ~ & T-GS \cite{luthermalgaussian} & 17.99 & 19.88 & 22.12 & 19.32 & 25.06 & 21.33 & 21.04 & 23.37 & 27.50 & 23.53 & 22.84 & 18.81 & 21.90 \\
        ~ & ~ & MMOne \cite{gu2025mmone} & 18.05 & 20.33 & 22.35 & 21.01 & 25.29 & 22.43 & 20.76 & 23.55 & 27.83 & 23.86 & 23.97 & 19.06 & 22.37 \\
        ~ & ~ & Ours & \textbf{26.45} & \textbf{33.15} & \textbf{32.83} & \textbf{30.20} & \textbf{32.47} & \textbf{28.26} & \textbf{30.83} & \textbf{32.27} & \textbf{34.57} & \textbf{33.96} & \textbf{38.14} & \textbf{33.73} & \textbf{32.24}\\
    \midrule[1pt]
        \multirow{6}{*}{RGB} & \multirow{6}{*}{PSNR$\uparrow$} & D-3DGS \cite{yang2024deformable} & 18.11 & 22.73& 22.36 & 18.32 & 15.08 & 14.35 & 27.10 & 16.80 & 22.95 & 16.45 & 15.52 & 18.17 & 18.99 \\
        ~ & ~ & 4DGaussians \cite{wu20244d} & 19.87 & 27.19 & 26.23 & 20.30 & 18.03 & 15.50 & 30.09 & 21.99 & 24.72 & 19.07 & 19.02 & 23.74 & 22.15  \\
        ~ & ~ & E-D3DGS \cite{bae2024per} & 21.55 & 27.68 & 28.00 & 23.25 & 18.60 & 15.66 & 31.73 & 23.75 & 26.84 & 22.39 & 22.45 & 25.38 & 23.94\\
        ~ & ~ & T-GS \cite{luthermalgaussian} & 15.96 & 16.61 & 18.81 & 11.89 & 11.43 & 13.20 & 21.64 & 13.49 & 14.26 & 11.70 & 12.58 & 15.57 & 14.76\\
        ~ & ~ & MMOne \cite{gu2025mmone} & 16.14 & 16.96 & 18.42 & 12.45 & 11.63 & 12.69 & 21.34 & 13.35 & 14.43 & 11.79 & 12.65 & 15.29 & 14.76\\
        ~ & ~ & Ours & \textbf{21.84} & \textbf{28.33} & \textbf{28.90} & \textbf{24.22} & \textbf{19.94} & \textbf{17.92} & \textbf{31.96} & \textbf{25.18} & \textbf{28.15} & \textbf{23.42} & \textbf{23.70} & \textbf{26.79} & \textbf{25.03}\\
    \bottomrule[2pt]
    \end{tabular}}
  \label{tab:quant}
\end{table*}

\begin{table}[!t]
	\caption{Average quantitative comparison across all scenes.}
	\centering
    \resizebox{1.0\linewidth}{!}{
        \begin{tabular}{lcccccc}  
	\toprule[1pt]
    \multirow{2}{*}{Methods} & \multicolumn{3}{c}{RGB} & \multicolumn{3}{c}{Thermal}\\
    \cmidrule(lr){2-4}\cmidrule(lr){5-7}
    ~ & PSNR$\uparrow$ & SSIM$\uparrow$ & LPIPS$\downarrow$ & PSNR$\uparrow$ & SSIM$\uparrow$ & LPIPS$\downarrow$ \\
    \midrule[1pt]
    \multirow{2}{*}{D-3DGS~\cite{yang2024deformable}} & 18.99 & 0.690 & 0.351 & -- & -- & -- \\
    ~ & -- & -- & -- & 25.65 & 0.916 & 0.170 \\
    \cmidrule(lr){2-7}
    \multirow{2}{*}{4DGaussians~\cite{wu20244d}} & 22.15 & 0.744 & 0.303 & -- & -- & -- \\
    ~ & -- & -- & -- & 30.24 & 0.930 & 0.148 \\
    \cmidrule(lr){2-7}
    \multirow{2}{*}{E-D3DGS~\cite{bae2024per}} & 23.94 & 0.795 & 0.238 & -- & -- & -- \\
    ~ & -- & -- & -- & 30.72 & 0.934 & 0.144 \\
    \cmidrule(lr){2-7}
    ThermalGaussian~\cite{luthermalgaussian} & 14.76 & 0.479 & 0.498 & 21.90 & 0.874 & 0.255 \\
    MMOne~\cite{gu2025mmone} & 14.76 & 0.491 & 0.494 & 22.37 & 0.886 & 0.253 \\
    \midrule[0.5pt]
    Ours & \textbf{25.03} & \textbf{0.828} & \textbf{0.206} & \textbf{32.24} & \textbf{0.947} & \textbf{0.119} \\
	\bottomrule[1pt]
	\end{tabular} 
    }
	\label{tab2:average_quant}
\end{table}

\section{Experiments}

\subsection{Implementation Details}
Our method builds upon E-D3DGS~\cite{bae2024per} with the same optimization settings. Training starts from RGB-based sparse points and camera poses shared with synchronized thermal frames. Each iteration first updates geometry and RGB attributes in the RGB deformation stage, then geometry and thermal attributes under thermal supervision. All Gaussians remain shared for 1,000 iterations before the routing logits become learnable. Specialization starts after 10,000 iterations; at 20,000 iterations, routing assignments are fixed and densification stops. We set the temperature $\tau$ to 1.0 and the routing-logit learning rate to 0.001. At inference, the target timestamp drives both deformation stages, routing masks select the Gaussians for each modality, and rendering from the target pose produces RGB and thermal outputs.
All methods are trained for 30,000 iterations on a single NVIDIA A5000 GPU. The resolution of both RGB and thermal images is 320$\times$240.

\subsection{Metrics and Baselines}
We evaluate RGB and thermal reconstruction using Peak Signal-to-Noise Ratio (PSNR)~\cite{hore2010image}, Structural Similarity (SSIM)~\cite{wang2004image}, and Learned Perceptual Image Patch Similarity (LPIPS)~\cite{zhang2018unreasonable}.

We compare two representative baseline groups. ThermalGaussian~\cite{luthermalgaussian} and MMOne~\cite{gu2025mmone} are static multimodal methods that jointly model RGB and thermal observations but do not handle dynamic scenes. D-3DGS~\cite{yang2024deformable}, 4DGaussians~\cite{wu20244d}, and E-D3DGS~\cite{bae2024per} are single-modality dynamic methods that model temporal variation without multimodal information. These groups cover static RGB-Thermal reconstruction and dynamic RGB reconstruction, providing relevant comparisons for our joint dynamic RGB-Thermal setting.

Since thermal imagery lacks stable texture for COLMAP-based pose estimation, we use RGB-derived sparse points and camera poses~\cite{schoenberger2016sfm} as the shared geometric reference. All methods thus use a consistent protocol, with results reported separately by modality.

\begin{figure*}[!t]
  \captionsetup[subfloat]{labelformat=empty}
	\centering
	\subfloat[ThermalGaussian]{
		\includegraphics[width=0.135\linewidth]{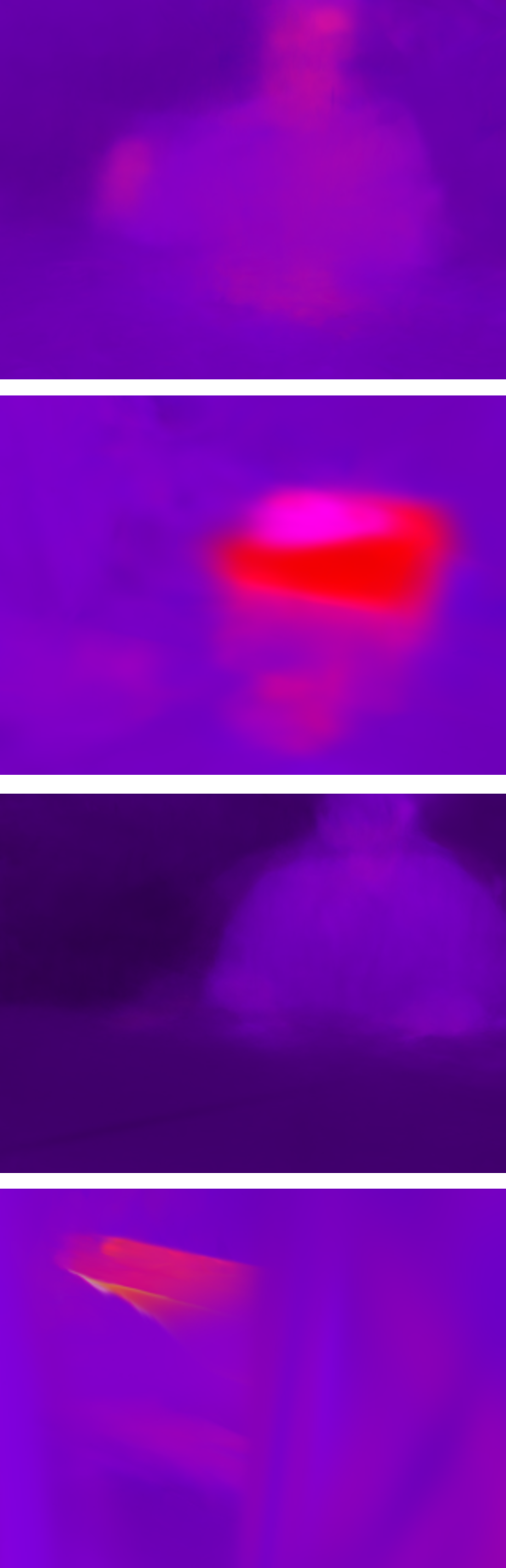}}
	\subfloat[MMOne]{
		\includegraphics[width=0.135\linewidth]{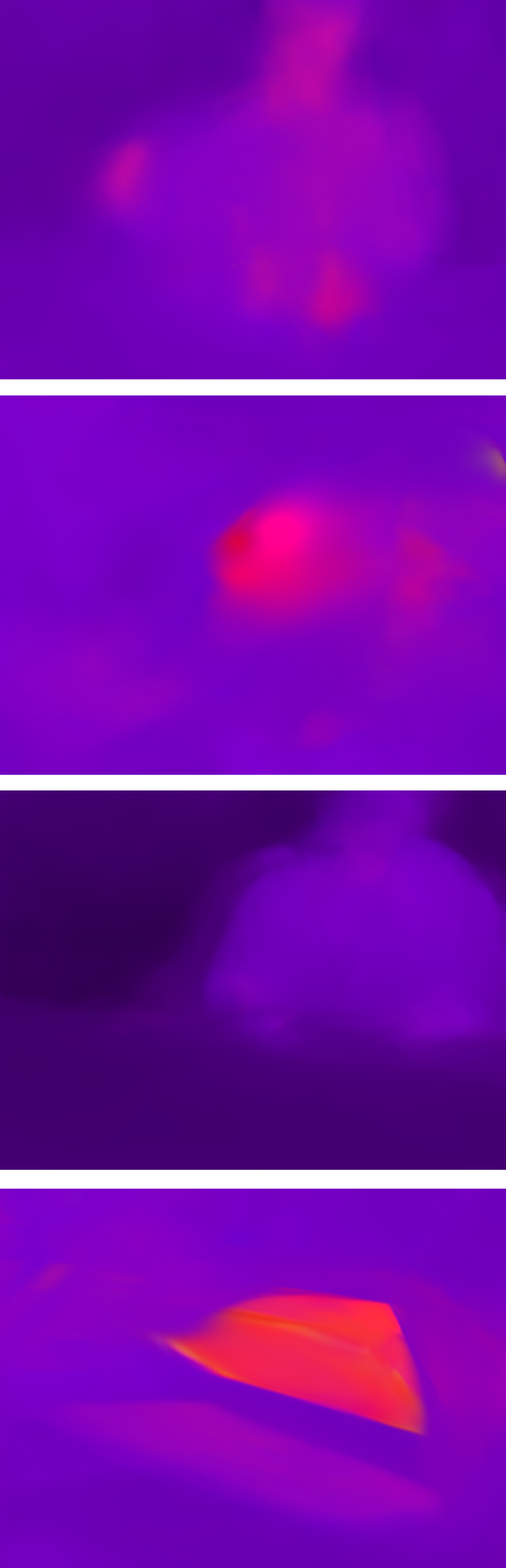}}
	\subfloat[D-3DGS]{
		\includegraphics[width=0.135\linewidth]{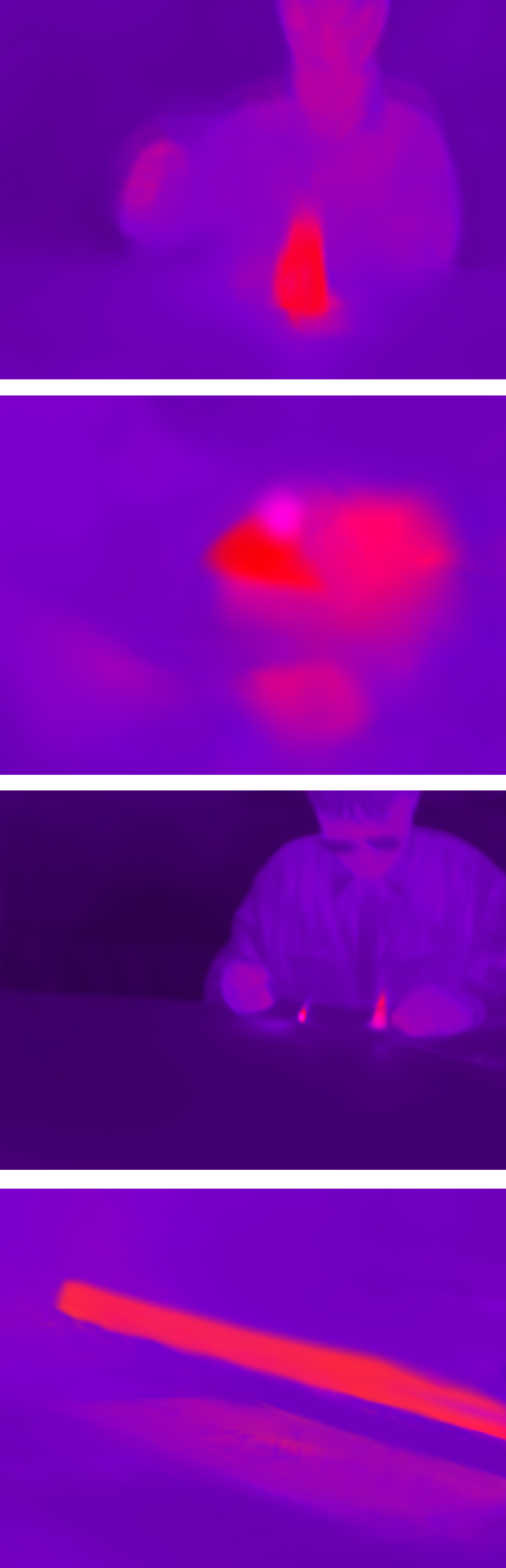}}
	\subfloat[4DGaussians]{
		\includegraphics[width=0.135\linewidth]{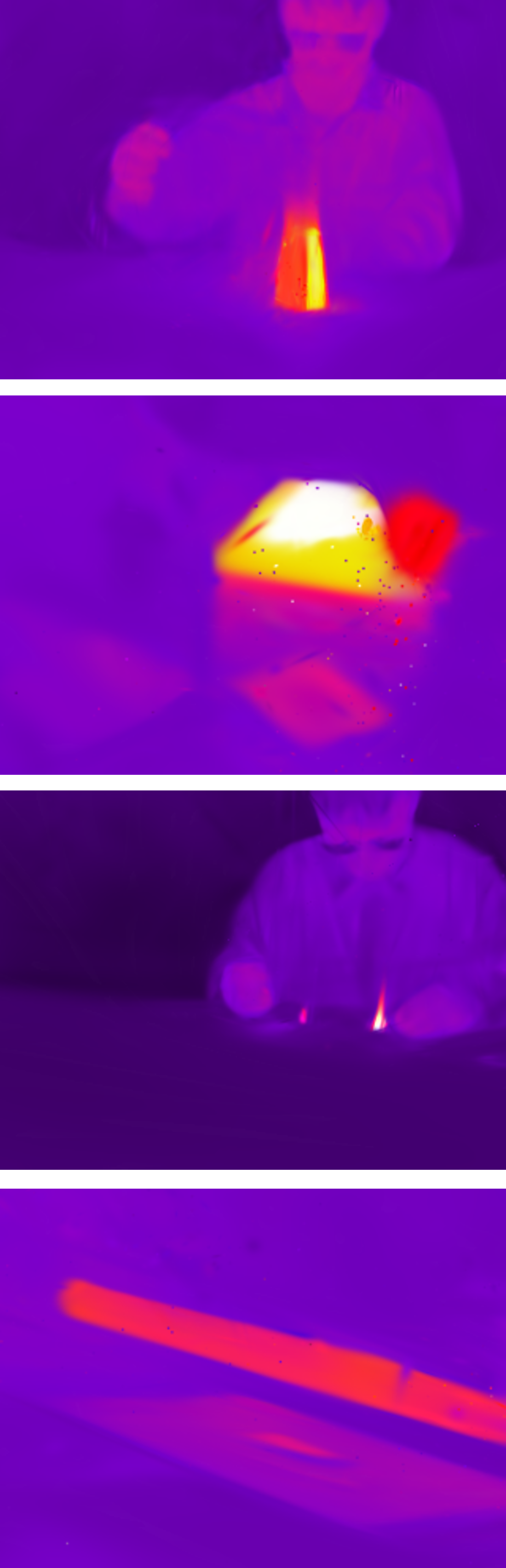}}	
	\subfloat[E-D3DGS]{
		\includegraphics[width=0.135\linewidth]{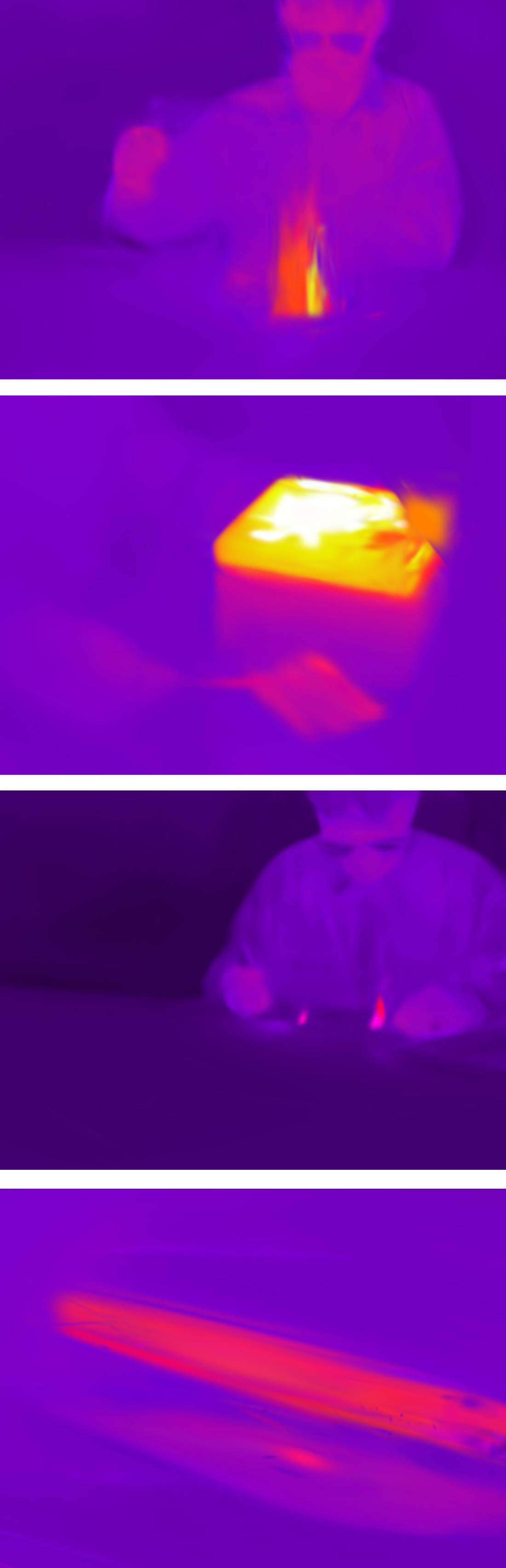}}
	\subfloat[Ours]{
		\includegraphics[width=0.135\linewidth]{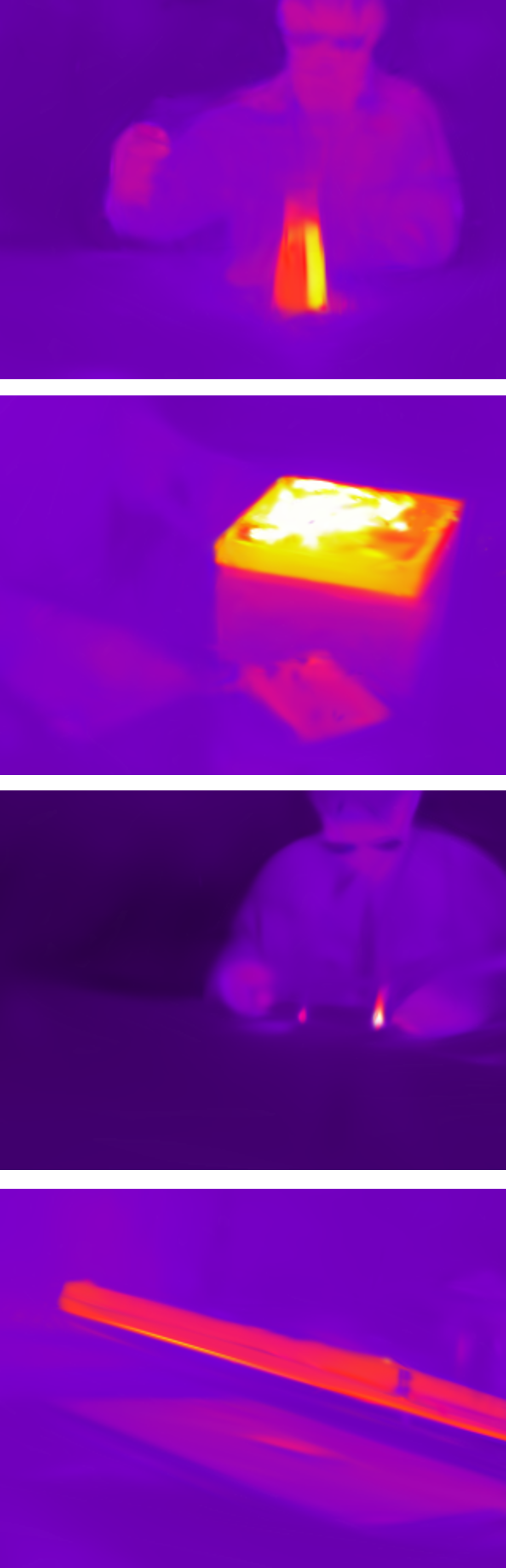}}
	\subfloat[GT]{
		\includegraphics[width=0.135\linewidth]{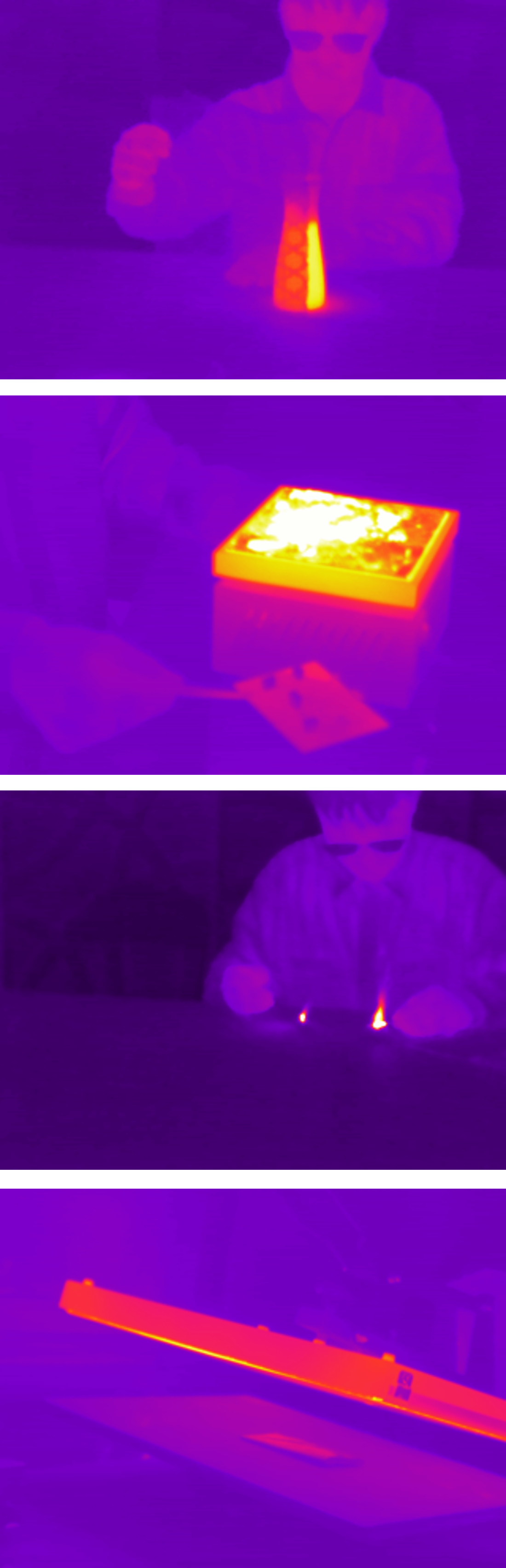}}
	\caption{Qualitative comparison of methods for thermal novel-view synthesis}
	\label{Fig5.2: thermal}	
\end{figure*}

\begin{figure*}[!t]
	\centering
  \setlength{\tabcolsep}{1pt}
  \begin{tabular}{>{\small}ccccccc}
  \makecell{Thermal-\\Gaussian}
  & \raisebox{-0.5\height}{\includegraphics[width=0.144\linewidth]{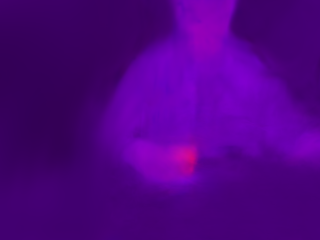}}
  & \raisebox{-0.5\height}{\includegraphics[width=0.144\linewidth]{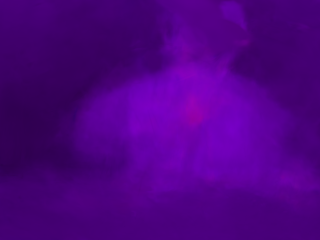}}
  & \raisebox{-0.5\height}{\includegraphics[width=0.144\linewidth]{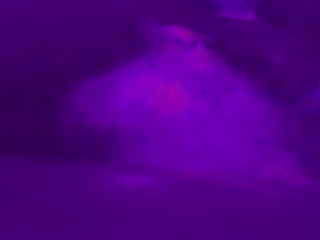}}
  & \raisebox{-0.5\height}{\includegraphics[width=0.144\linewidth]{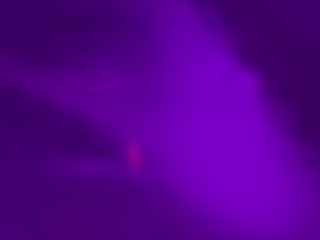}}
  & \raisebox{-0.5\height}{\includegraphics[width=0.144\linewidth]{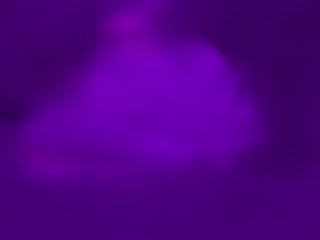}}
  & \raisebox{-0.5\height}{\includegraphics[width=0.144\linewidth]{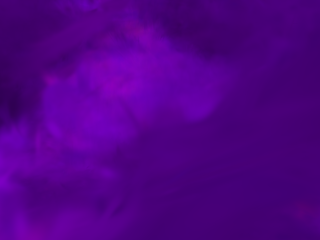}} \rule{0pt}{1cm}\\

  E-D3DGS
  & \raisebox{-0.5\height}{\includegraphics[width=0.144\linewidth]{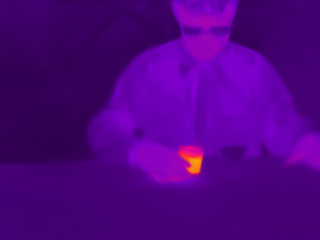}}
  & \raisebox{-0.5\height}{\includegraphics[width=0.144\linewidth]{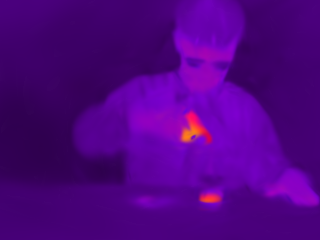}}
  & \raisebox{-0.5\height}{\includegraphics[width=0.144\linewidth]{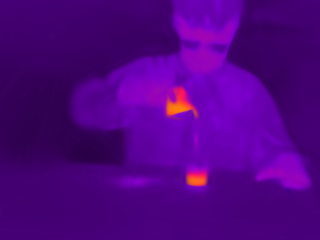}}
  & \raisebox{-0.5\height}{\includegraphics[width=0.144\linewidth]{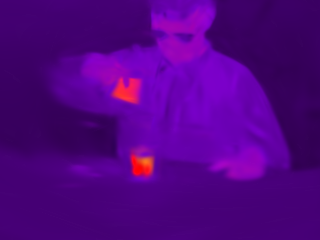}}
  & \raisebox{-0.5\height}{\includegraphics[width=0.144\linewidth]{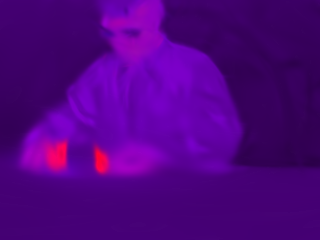}}
  & \raisebox{-0.5\height}{\includegraphics[width=0.144\linewidth]{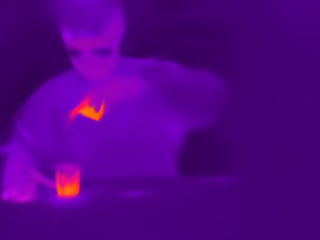}} \rule{0pt}{1cm} \\

  Ours
  & \raisebox{-0.5\height}{\includegraphics[width=0.144\linewidth]{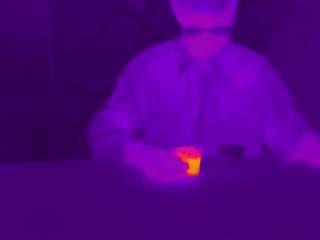}}
  & \raisebox{-0.5\height}{\includegraphics[width=0.144\linewidth]{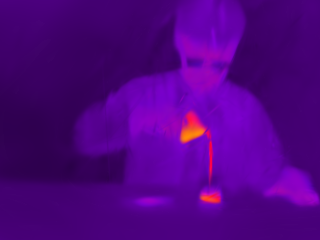}}
  & \raisebox{-0.5\height}{\includegraphics[width=0.144\linewidth]{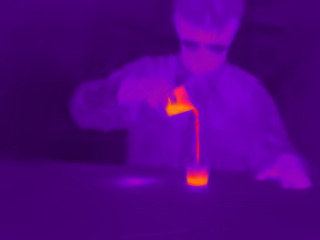}}
  & \raisebox{-0.5\height}{\includegraphics[width=0.144\linewidth]{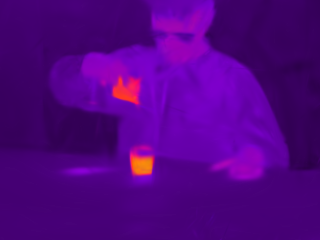}}
  & \raisebox{-0.5\height}{\includegraphics[width=0.144\linewidth]{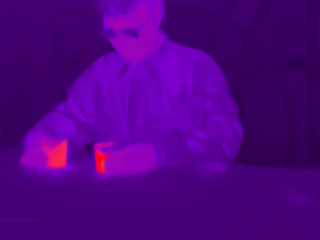}}
  & \raisebox{-0.5\height}{\includegraphics[width=0.144\linewidth]{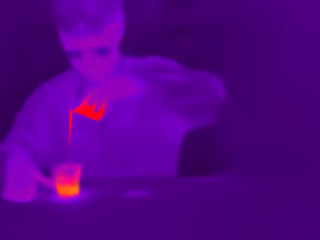}} \rule{0pt}{1cm} \\

  GT
  & \raisebox{-0.5\height}{\includegraphics[width=0.144\linewidth]{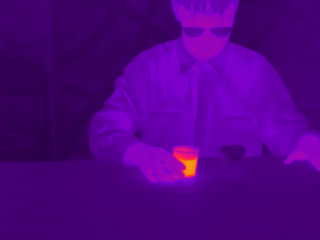}}
  & \raisebox{-0.5\height}{\includegraphics[width=0.144\linewidth]{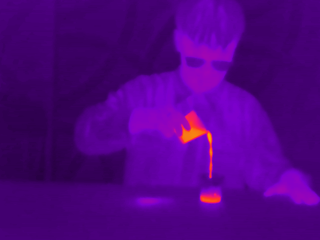}}
  & \raisebox{-0.5\height}{\includegraphics[width=0.144\linewidth]{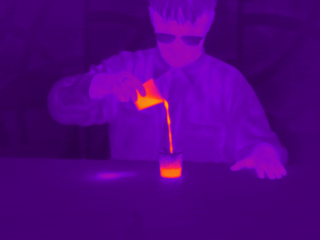}}
  & \raisebox{-0.5\height}{\includegraphics[width=0.144\linewidth]{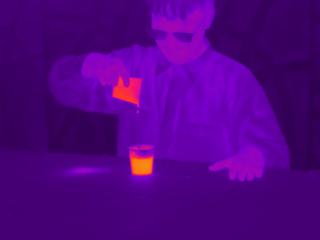}}
  & \raisebox{-0.5\height}{\includegraphics[width=0.144\linewidth]{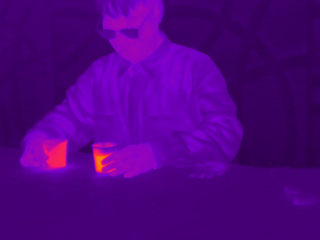}}
  & \raisebox{-0.5\height}{\includegraphics[width=0.144\linewidth]{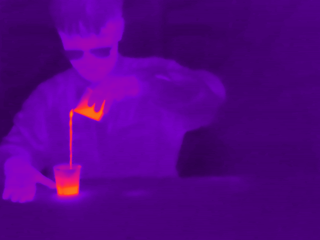}} \rule{0pt}{1cm} \\

   & \#26 & \#153 & \#197 & \#308 & \#418 & \#583 \\
  \end{tabular}
   \caption{Qualitative comparison across timesteps for representative methods in \textit{HotWater}.}
\label{fig5.3:thermal-time}
\end{figure*}

\subsection{Thermal View Synthesis}

Tables~\ref{tab:quant} and~\ref{tab2:average_quant} report thermal results. Our method achieves the best averages of 32.24 dB PSNR, 0.947 SSIM, and 0.119 LPIPS, exceeding the strongest dynamic baseline by 1.52 dB in PSNR. In dynamic scenes, static multimodal methods fit mixed temporal states rather than reconstructing a time-varying thermal field. Single-modality dynamic methods model temporal variation without complementary cross-modal information. Our method instead combines both modalities in a unified dynamic process and renders high-fidelity RGB and thermal images at arbitrary timesteps and viewpoints.

Figure~\ref{Fig5.2: thermal} compares novel-view thermal rendering across scenes. Static multimodal methods recover only blurred scene outlines in dynamic settings. Single-modality dynamic methods improve dynamic detail recovery but still show clear distortions in high-temperature or rapidly varying regions because they lack multimodal information. In contrast, our renderings remain closer to the ground truth, preserve clearer structures in high-temperature areas, and show more consistent temperature transitions.
Figure~\ref{fig5.3:thermal-time} further compares representative timesteps, providing evidence of temporal consistency under rapidly changing thermal patterns. ThermalGaussian performs poorly across timesteps, showing that a static representation cannot handle multimodal dynamic reconstruction. E-D3DGS recovers overall dynamic motion but misses fine-grained details in rapidly changing regions because it lacks multimodal modeling, such as the stream of poured hot water. Our method captures both global motion and local detail under fast temperature changes.

\begin{figure*}[!t]  
	\centering
	\subfloat[ThermalGaussian]{
		\includegraphics[width=0.2\linewidth]{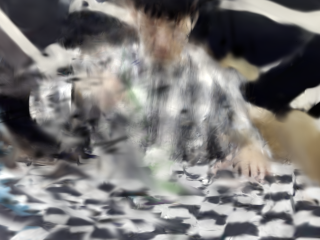}
	}	
		\subfloat[MMOne]{
		\includegraphics[width=0.2\linewidth]{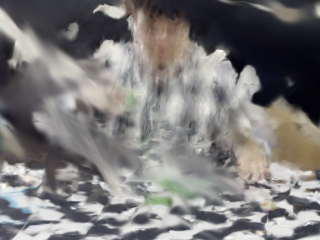}
	}	
	\subfloat[D-3DGS]{
		\includegraphics[width=0.2\linewidth]{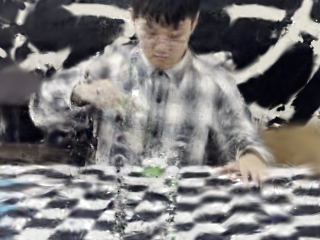}
	}	
	\subfloat[4DGaussians]{
		\includegraphics[width=0.2\linewidth]{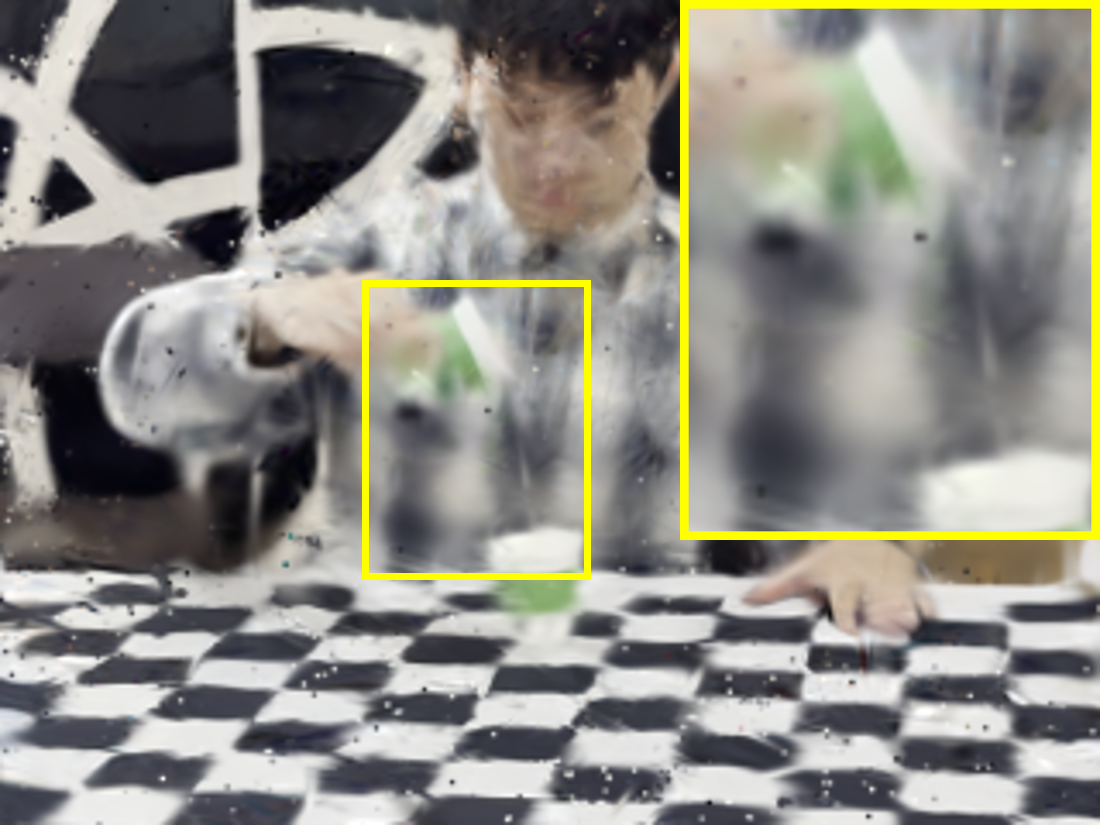}
	}	\\
	\subfloat[E-D3DGS]{
		\includegraphics[width=0.2\linewidth]{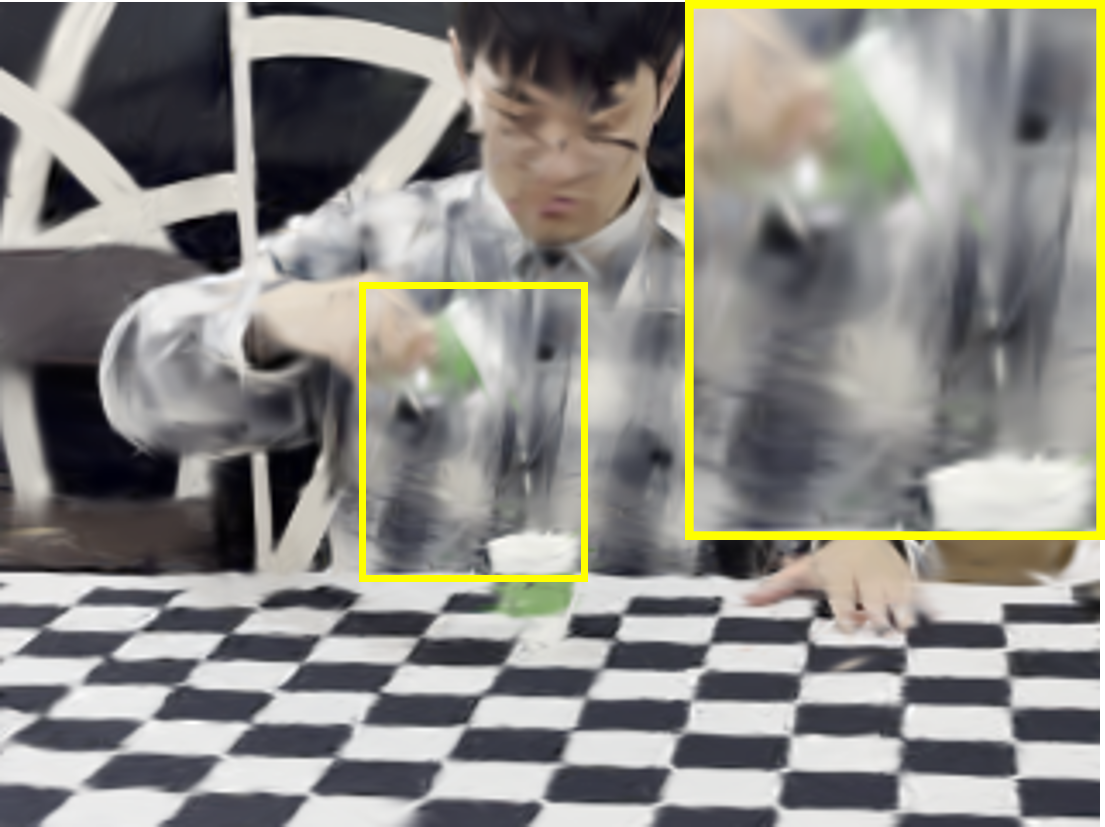}
	}
	\subfloat[Ours]{
		\includegraphics[width=0.2\linewidth]{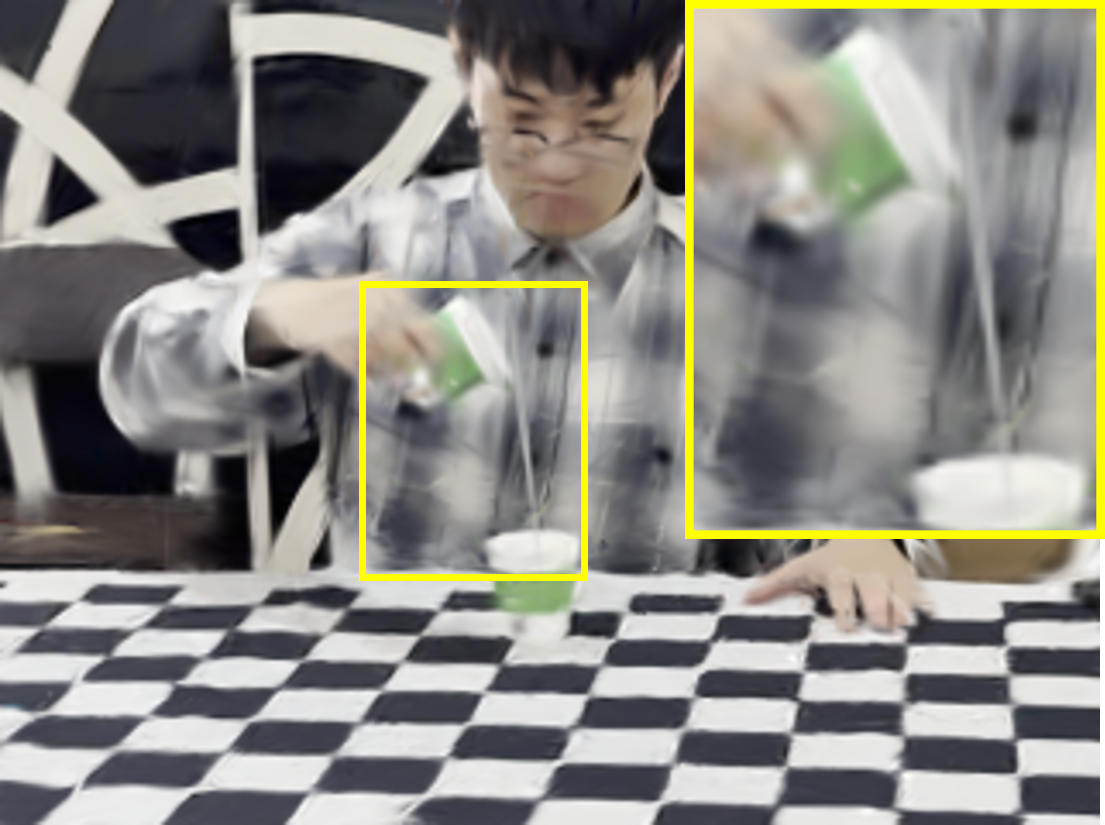}
	}
	\subfloat[RGB GT]{
		\includegraphics[width=0.2\linewidth]{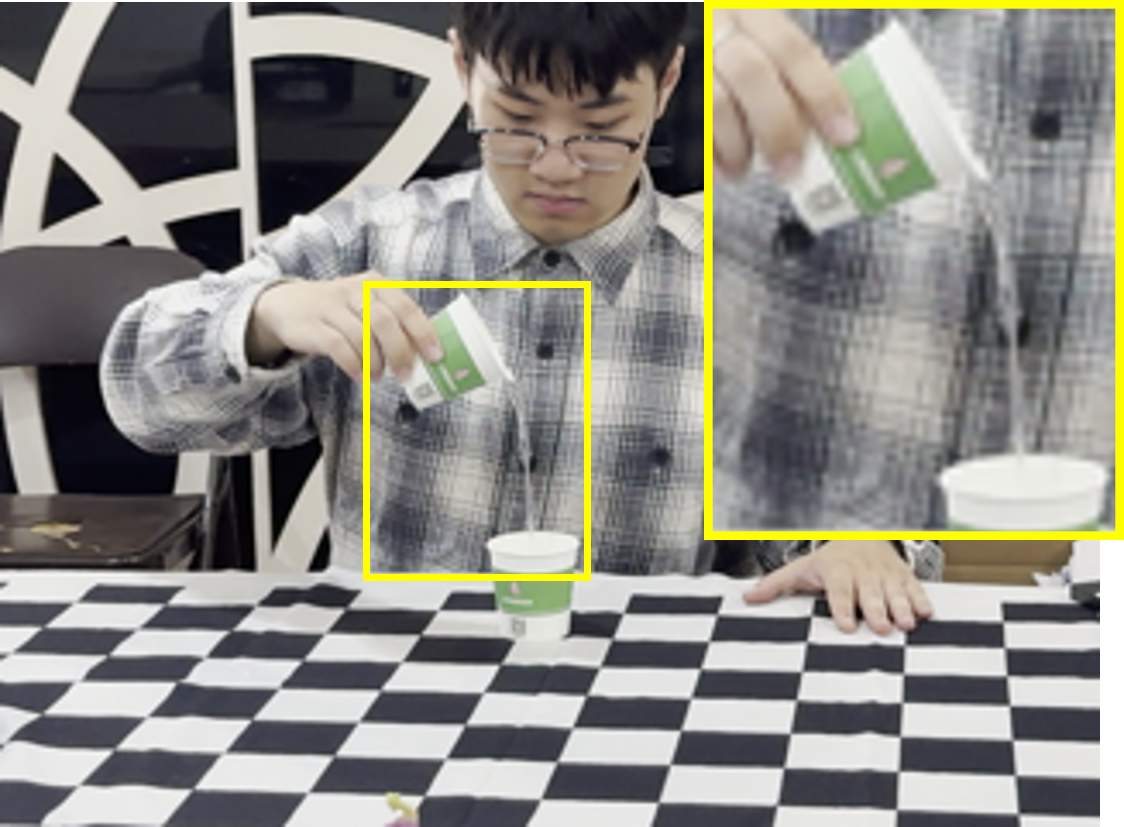}
	}
	\subfloat[Thermal GT]{
		\includegraphics[width=0.2\linewidth]{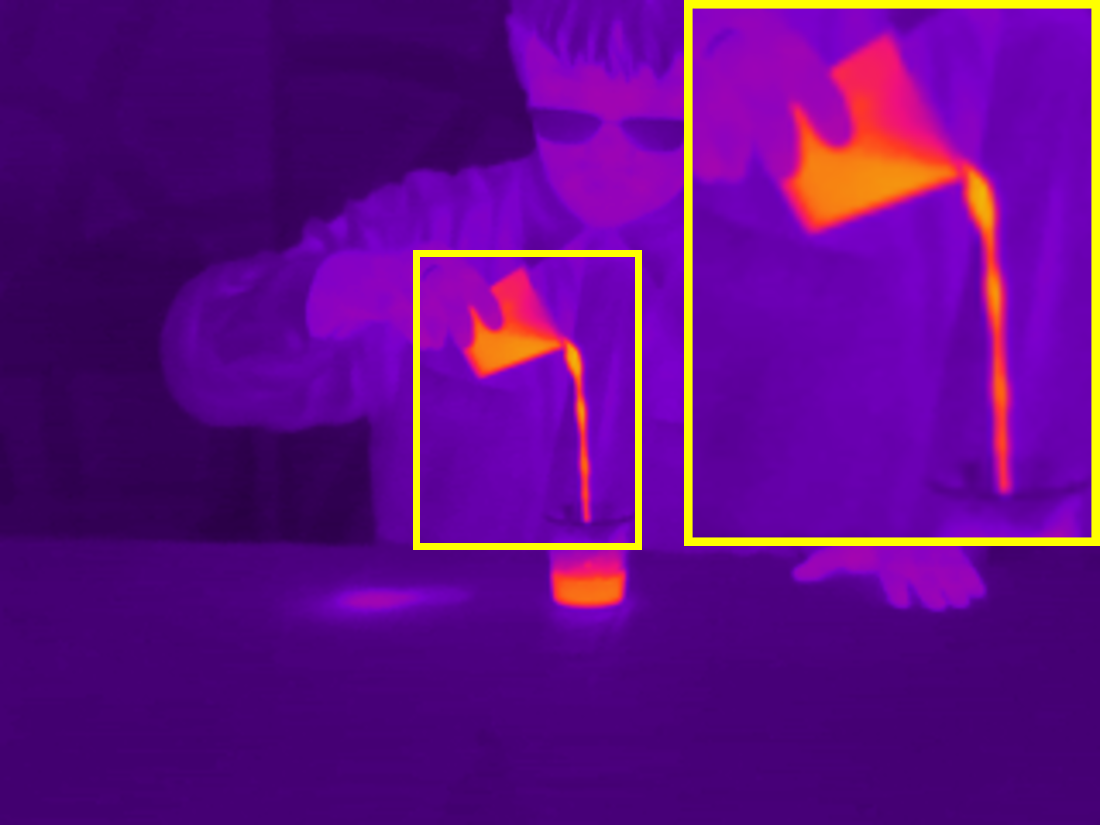}
	}
	\caption{Qualitative comparison of different methods for RGB novel-view synthesis.
    }
	\label{fig5.4:rgb_qualitative}	
\end{figure*}

\subsection{RGB View Synthesis}

Multimodal supervision improves not only thermal reconstruction but also RGB dynamic reconstruction. Table~\ref{tab:quant} shows that our method achieves the best RGB averages, with 25.03 dB PSNR, 0.828 SSIM, and 0.206 LPIPS, exceeding the strongest baseline by 1.09 dB in PSNR. Static multimodal methods degrade substantially on this task because they lack dynamic modeling. Although D-3DGS, 4DGaussians, and E-D3DGS are designed for dynamic RGB reconstruction, our method outperforms them across all scenes, showing that cross-modal optimization also benefits RGB reconstruction.

Figure~\ref{fig5.4:rgb_qualitative} compares RGB renderings. The static methods ThermalGaussian and MMOne show clear motion blur and geometric distortion. D-3DGS improves global geometry, but high-frequency details remain blurry. 4DGaussians and E-D3DGS further improve geometric completeness and approach the ground truth in static backgrounds, yet still lose detail in the fast-moving, low-contrast regions marked by yellow boxes, such as flowing water. This stream has strong radiative contrast against the background in thermal images. Multimodal supervision therefore helps our method preserve dynamic detail and produce results closer to the ground truth in these challenging regions. These results further validate our method for dynamic RGB novel-view synthesis.

\begin{table}[!t]
	\caption{Ablation study results averaged over all scenes.}
	\centering
    \resizebox{1.0\linewidth}{!}{
        \begin{tabular}{lcccccc}  
	\toprule[1pt]
    \multirow{2}{*}{Methods} & \multicolumn{3}{c}{RGB} & \multicolumn{3}{c}{Thermal}\\
    \cmidrule(lr){2-4}\cmidrule(lr){5-7}
    ~ & PSNR$\uparrow$ & SSIM$\uparrow$ & LPIPS$\downarrow$ & PSNR$\uparrow$ & SSIM$\uparrow$ & LPIPS$\downarrow$ \\
    \midrule[1pt]
    E-D3DGS(RGB) & 23.94 & 0.795 & 0.238 & -- & -- & -- \\
    E-D3DGS(Thermal) & -- & -- & -- & 30.72 & 0.934 & 0.144 \\
    + shared Gaussian & 24.58 & 0.816 & 0.220 & 30.23 & 0.931 & 0.144 \\
    + multimodal embeddings & 24.89 & 0.826 & 0.208 & 30.72 & 0.936 & 0.139\\
    + multimodal routing & \textbf{25.03} & \textbf{0.828} & \textbf{0.206} & \textbf{32.24} & \textbf{0.947} & \textbf{0.119} \\
	\bottomrule[1pt]
	\end{tabular} 
    }
	\label{tab3:ablation}
\end{table}

\begin{figure}[!t]  
  \centering
  \subfloat[Shared Gaussians (RGB)]{
    \includegraphics[width=0.45\linewidth]{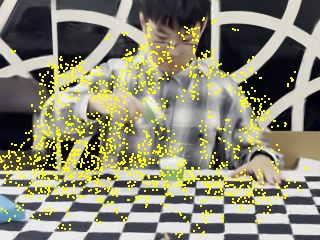}}%
  \hfill
  \subfloat[RGB-only Gaussians]{
    \includegraphics[width=0.45\linewidth]{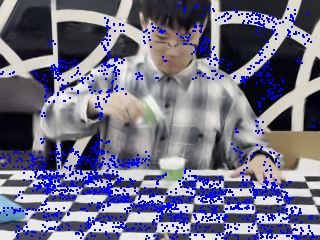}}\\

  \subfloat[Shared Gaussians (Thermal)]{
    \includegraphics[width=0.45\linewidth]{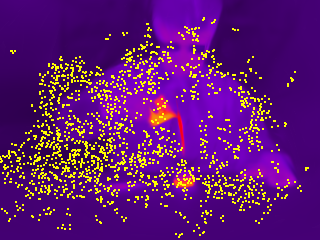}}%
  \hfill
  \subfloat[Thermal-only Gaussians]{
    \includegraphics[width=0.45\linewidth]{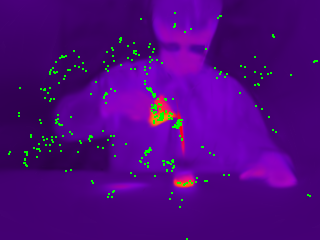}}
  \caption{
  Spatial distribution of different Gaussians in RGB and thermal views after multimodal routing. 
  }
  \label{fig:pointcloud}	
\end{figure}

\subsection{Ablation Studies}

We conduct ablation studies on DynamicRGBT-Scenes to evaluate the contribution of each component. Table~\ref{tab3:ablation} reports results for the RGB and thermal modalities, using E-D3DGS(RGB/Thermal) as the corresponding baselines.

Starting from the modality-specific baselines, the shared multimodal Gaussian representation enables joint dynamic reconstruction of both modalities within a unified framework. This design improves RGB performance but slightly reduces thermal reconstruction quality. We then add multimodal embeddings to disentangle temporal codes and Gaussian embeddings across modalities. Both modalities improve, with thermal performance nearly matching its single-modality baseline. This result shows that the embeddings reduce feature coupling between heterogeneous modalities during joint modeling. Finally, multimodal routing consistently improves both modalities; for thermal reconstruction, it yields a clear PSNR gain of 1.52 dB.

To analyze the spatial distribution induced by multimodal routing, Figure~\ref{fig:pointcloud} projects sampled Gaussians from the HotWater scene onto RGB and thermal images. Shared Gaussians concentrate in regions with correlated cross-modal variation. RGB-only Gaussians mainly cover textured backgrounds with limited temperature variation and also cover some moving regions to preserve visual detail. In contrast, thermal-only Gaussians concentrate in regions with significant temperature changes, including hot water, flowing streams, and their boundaries. These observations show that multimodal routing effectively separates shared and modality-specific Gaussians, improving each modality's capacity to model fine-grained regions. Overall, every component contributes positively; together, these gains support the effectiveness of the proposed multimodal dynamic reconstruction framework.

\section{Conclusions}
In this work, we present DynamicRGBT-Scenes, a benchmark for joint dynamic RGB-Thermal reconstruction, and Dynamic Thermal Gaussians, the first dynamic RGB-Thermal reconstruction framework for complex scenes. The framework jointly models RGB and thermal modalities through a shared geometric representation, modality embeddings, and a structure-aware routing mechanism. Combined with a dynamic deformation strategy, it effectively decouples geometric motion from modality-specific variations, enabling stable reconstruction of rapidly evolving thermal fields and supporting novel-view synthesis at arbitrary timesteps and viewpoints.

The dataset's synchronized high-frequency observations support 4D thermal analysis. Extensive experiments show state-of-the-art reconstruction performance and confirm the benefits of multimodal embeddings and adaptive routing for spatiotemporal thermography in real-world environments.

\clearpage
\begin{acks}
This work was supported by the National Key Research and Development Program of China (No. 2023YFB4502803), the National Natural Science Foundation of China (No. 12371349), and the Joint Fund of Zhejiang Provincial Natural Science Foundation of China (No. LLSSZ25F030002).
\end{acks}

\bibliographystyle{ACM-Reference-Format}
\bibliography{sample-base}

\end{document}